\documentclass[letterpaper, 10 pt, conference]{ieeeconf}  

\IEEEoverridecommandlockouts                              

\usepackage{graphics} 
\usepackage{epsfig} 
\usepackage{caption,subcaption}
\usepackage{amsmath}
\usepackage{eufrak}
\usepackage{booktabs}
\usepackage{tabularx}
\usepackage{graphicx}
\usepackage{sidecap}
\usepackage{todonotes}
\sidecaptionvpos{figure}{c}
\usepackage{stackengine}
\usepackage{cite}

\newcommand\inv[1]{#1\raisebox{1.15ex}{$\scriptscriptstyle-\!1$}}

\def\Vec#1{\!\!\hbox{$#1$\kern-0.38em\lower0.85em\hbox{$\vec{}\,$}}\,}%
\newcommand{\bbm}{\begin{bmatrix}}
\newcommand{\ebm}{\end{bmatrix}}

\usepackage{acronym}
\acrodef{ARM}{Advanced RISC (Reduced Instruction Set Computer) Machine}
\acrodef{AGL}{Above Ground Level}
\acrodef{uav}[UAV]{Unmanned Aerial Vehicle}
\acrodef{vtr}[VT\&R]{Visual Teach and Repeat}
\acrodef{GPS}{Global Positioning System}
\acrodef{IMU}{Inertial Measurement Unit}
\acrodef{LiDAR}{Light Detection And Ranging}
\acrodef{MET}{Mars Emulation Terrain}
\acrodef{MATS}{Multi-Agent Tactical Sentry}
\acrodef{MAV}{Micro Aerial Vehicle}
\acrodef{MLESAC}{Maximum Likelihood Estimation SAmple Consensus}
\acrodef{PGR}{Point Grey Research}
\acrodef{RANSAC}{RANdom SAmple Consensus}
\acrodef{GRIC}{Geometric Robust Information Criterion}
\acrodef{SLAM}{Simultaneous Localisation and Mapping}
\acrodef{STPG}{Spatio-Temporal Pose Graph}
\acrodef{EBN}{Experience-Based Navigation}
\acrodef{STEAM}{Simultaneous Trajectory Estimation And Mapping}
\acrodef{MEL}{Multi-Experience Localisation}
\acrodef{SVM}{Support Vector Machine}
\acrodef{MPC}{Model Predictive Control}
\acrodef{PnP}{Perspective-Three-Point}
\acrodef{PTAM}{Parallel Tracking and Mapping}
\acrodef{SURF}{Speeded-Up Robust Features}
\acrodef{VO}{Visual Odometry}
\acrodef{cdf}[CDF]{Cumulative Distribution Function}
\acrodef{RUV}{Robotic Utility Vehicle}
\acrodef{FOV}{Field Of View}
\acrodef{VINS}{Visual-Inertial Navigation Systems}
\acrodef{ILC}{Iterative Learning Control}
\acrodef{RUV}{Robotic Utility Vehicle}
\acrodef{GPU}{Graphics Processing Unit}
\acrodef{DoF}{Degree of Freedom}
\acrodef{CDF}{Cumulative Distribution Function}
\acrodef{UTIAS}{University of Toronto Institute for Aerospace Studies}
\acrodef{BVLOS}{Beyond Visual Line-Of-Sight}
\acrodef{VLOS}{Visual Line-Of-Sight}
\acrodef{GNSS}{Global Navigation Satellite Systems}
\acrodef{DRDC}{Defence Research \& Development Canada}
\acrodef{USB}{Universal Serial Bus}
\acrodef{TTL}{Transistor-Transistor Logic}
\acrodef{RAM}{Random Access Memory}
\acrodef{ROS}{Robot Operating System}

\title{\LARGE \bf
There's No Place Like Home: Visual Teach and Repeat for Emergency Return of Multirotor UAVs During GPS Failure}

\author{Michael Warren, Melissa Greeff, Bhavit Patel, Jack Collier, Angela P. Schoellig, and Timothy D. Barfoot
\vspace{-8mm}
\thanks{Jack Collier is with Defence Research and Development Canada: Jack.Collier@drdc-rddc.gc.ca.  All other authors are with the University of Toronto Institute for Aerospace Studies (UTIAS), University of Toronto, Canada: \{michaelwarren, melissa.greeff, bhavit.patel\}@robotics.utias.utoronto.ca, \{angela.schoellig, tim.barfoot\}@utoronto.ca. Accompanying video available at: \textbf{tiny.cc/noplacelikehome}}%
}

\begin{document}
\maketitle
\thispagestyle{empty}
\pagestyle{empty}

\vspace{-5mm}
\begin{abstract}
Redundant navigation systems are critical for safe operation of \acsp{uav} in high-risk environments. Since most commercial \acsp{uav} almost wholly rely on \acs{GPS}, jamming, interference and multi-pathing are real concerns that usually limit their operations to low-risk environments and \acs{VLOS}. This paper presents a vision-based route-following system for the autonomous, safe return of \acsp{uav} under primary navigation failure such as \acs{GPS} jamming. Using a \acl{vtr} framework to build a visual map of the environment during an outbound flight, we show the autonomous return of the \acs{uav} by visually localising the live view to this map when a simulated \acs{GPS} failure occurs, controlling the vehicle to follow the safe outbound path back to the launch point. Using gimbal-stabilised stereo vision alone, without reliance on external infrastructure or inertial sensing, \acl{VO} and localisation are achieved at altitudes of 5-25 m and flight speeds up to 55 km/h. We examine the performance of the visual localisation algorithm under a variety of conditions and also demonstrate closed-loop autonomy along a complicated 450 m path.
\end{abstract}

\vspace{-2mm}
\section{INTRODUCTION}\label{sec:introduction}
Safe beyond \ac{VLOS} operations are critical to enhancing the utility of \acp{uav} in large-scale, outdoor operations. Typically, reliance on \ac{GNSS} for navigation in most low-cost commercial \acp{uav} mean the authorisation to do so from government regulators is rare. Jamming, interference and accuracy concerns mean that \ac{GPS} alone cannot be relied on in cases of close-proximity, safety-critical or high-value operations. In this paper, we present a complete vision-only route-following system for the autonomous navigation of \acp{uav}, and demonstrate its use as a functional backup system for \ac{GPS}-only navigation. Using this system allows the vehicle to navigate home visually in case of primary navigation system failure, without reliance on any external infrastructure, or inertial sensing for the vision-based components.

\ac{vtr} is a path-following algorithm capable of autonomously driving a robot by following a previously traversed route \cite{Furgale2010d}. Using visual feature matches from a live view to a locally metric map of 3D points allows the robot to estimate a path offset and send corrections to a path-following controller \cite{Ostafew2013}. Traditionally, \ac{vtr} is used on wheeled vehicles \cite{Paton2016}, with applications over constrained paths where external navigation infrastructure is unreliable or not available, e.g., factory floors, orchards, mines, urban road networks, and exploratory search-and-return missions. Using \ac{vtr} on aerial platforms has a number of unique use cases: just-in-time deliveries between warehouses, where flight paths are generally restricted to a few, high-frequency routes; monitoring of sensitive assets such as property borders or high-value infrastructure; and autonomous patrol in close-proximity environments, where poor sky view and jamming are notable concerns. Significantly, we want the vehicle to be able to autonomously and safely return to the take-off location at any time by using vision to localise to a map generated during the outbound path, all during a single flight.

\begin{figure}
  \centering
  \includegraphics[width=0.49\textwidth]{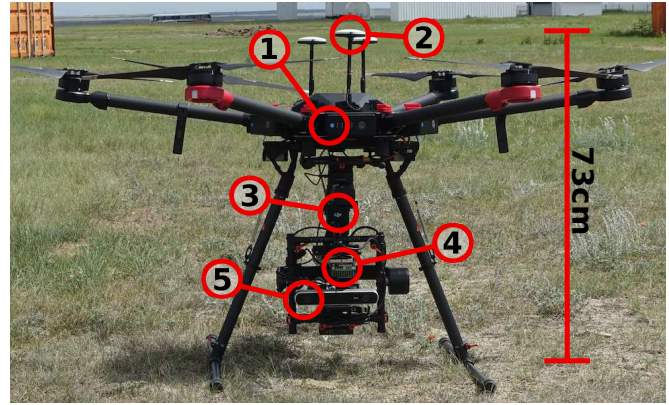}
  \caption{The experimental setup for  Visual Teach \& Repeat on our multirotor \ac{uav}: (1) DJI Matrice 600 Pro vehicle platform, (2) DJI A3 triple-redundant GPS module, (3) DJI Ronin-MX 3-axis gimbal, (4) NVIDIA Tegra TX2, (5) StereoLabs ZED camera.}
  \label{fig:m600}
  \vspace{-8mm}
\end{figure}

In this paper, we adapt the traditional \ac{vtr} methodology to suit these target use cases and apply our \ac{vtr} 2.0 system~\cite{Paton2016} on-board a multirotor \ac{uav} (Fig. \ref{fig:m600}) to demonstrate closed-loop operation. We show results of live localisation at speeds up to 15 m/s (55 km/h) at low altitude (5-25 metres) in winds up to 8 m/s, and demonstrate vision-based path-following control for the return segment of a just-taught outbound path. The novel work of this paper includes 1) demonstration of the \ac{vtr} framework on a new platform, a UAV with gimballed camera, 2) a thorough analysis of localisation performance and 3) presentation of a new path-following controller for multirotor \acp{uav}, all in a wide range of outdoor test scenarios. 

The rest of this paper is outlined as follows: Section \ref{sec:previous_work} examines similar work in visual route following for ground vehicles and \acp{uav}, and explores recent work in autonomous vision-based navigation of \acp{uav}. Section \ref{sec:methodology} describes the \ac{vtr} methodology for application on our target \ac{uav}, including the \ac{vtr} framework, localisation algorithm and gimbal and vehicle controllers. Section \ref{sec:experiments} describes the experimental setup to test the airborne \ac{vtr} framework, as well as description of datasets, field tests and results. The paper is concluded in Section \ref{sec:conclusion}.
\vspace{-4mm}
\section{PREVIOUS WORK}\label{sec:previous_work}
\vspace{-2mm}
\ac{vtr} and similar route-based navigation algorithms have a rich history on ground platforms \cite{Krajnik2010,Furgale2010d,Patona,Ostafew2013}, with the most recent extension adapted to include multiple experiences, increasing the autonomous performance time from a few days to several months \cite{Paton2016}. On \acp{uav}, there are now several demonstrations of teach-and-repeat style algorithms from the authors of this paper and others \cite{Pfrunder2014,Warren2017,Toudeshki2018,Surber2017}.

Our previous work, demonstrating the localisation performance of \ac{vtr} on fixed-wing \acp{uav} \cite{Warren2017} and integration of a gimballed camera on a ground vehicle \cite{Warren2018}, is the lead-up to this work. While there are few examples using a gimbaled camera on ground vehicles, a number of examples exist in demonstrations on \acp{uav} \cite{Playle2015,Sharp2001,Borowczyk2017,Lin2014,Choi2018}. This discrepency can most likely be attributed to the larger dynamic motions of \acp{uav}, where the utility of a gimbal is highly justified to ensure smooth sensor motion. In all the above cases, however, only two-axis gimbals are utilised. In our setup, we use an off-the-shelf three-axis gimbal to attenuate motion in all three rotational axes.

The approach that is closest conceptually to our work, with specific application on \acp{uav}, is \cite{Surber2017}. Despite not being framed as a `teach-and-repeat' technique, this system presents a demonstration of such a method on a \ac{uav} using a visual-inertial framework with weak GPS priors to assist initialisation of localisation and inform loop closures. Our work differs in that it requires no offline map building (the map is built on-board in real time) and does not require inertial sensors or external infrastructure such as \ac{GPS} for the perception component of the system.

Beyond the \ac{vtr} paradigm, there is a rich demonstration of vision-based navigation on \acp{uav} in recent years \cite{Giusti2015}. While most older demonstrations incorporate stereo camera systems for scale, they suffer from poor (i.e., small) baseline-to-depth ratios at higher altitudes. Recent advances in \ac{IMU} technology have allowed the use of loosely \cite{Achtelik2011,Weiss2011,Weiss2013b} and tightly coupled \cite{Hinzmann16,Forster2017,Mur-Artal2017} visual-inertial systems using both monocular and stereo cameras~\cite{Sun2017}, and with impressive demonstrations of dynamic maneuvers at high speed \cite{Loianno2017} in indoor, small scale setups. The majority of large scale demonstrations using these systems, however, often exist as a full \ac{SLAM} framework \cite{Shen2014,Qin}, incorporating exploration and globally metric 3D maps as a method of accurate survey. In contrast, \ac{vtr} takes a locally metric approach for map building, and leverages a human operator for the initial `demonstration' task, circumventing the difficult tasks of autonomous exploration and loop-closures.
\vspace{-2mm}
\section{METHODOLOGY}\label{sec:methodology}
In this paper, we use our well-established \ac{vtr} 2.0 software system as presented in \cite{Paton2016}, including the extension of a gimbaled camera \cite{Warren2018}. However, we adapt this system for use on a multirotor \ac{uav} specifically for the purposes of emergency return. Instead of \textit{teach} and \textit{repeat} phases, we implement functionally similar \textit{learn} and \textit{return} phases. 

During the \textit{learn} phase, the \ac{uav} flies using autonomous \ac{GPS} waypoint following or human operator control. During this phase, the \ac{vtr} algorithm performs passive \ac{VO}, inserting the visual observations from this privileged experience into a relative map of pose and scene structure, effectively `learning' the route. Following a primary navigation systems failure, the \ac{uav} should enter the \textit{return} phase, and, without reliance on \ac{GPS} or other external sensing, autonomously re-follow the route home in the reverse direction. In addition to performing  the same \ac{VO} as in \textit{learn}, it performs a localisation using a local segment from the learnt path. The vehicle follows the learnt path by sending high-frequency localisation updates (relative position and orientation with respect to the map) to a path-following controller. Once the vehicle returns to the start point, it hovers until taken over by a human controller. To be clear, in this paper we only simulate GPS failures, by manually commanding the vehicle to enter the return phase during flight.

In the following sections, we describe our \ac{vtr} system, including the architecture of the system, the visual navigation algorithm, and gimbal and path-following controllers.

\subsection{System Overview}
\begin{figure}
\vspace{2mm}
  \centering
  \includegraphics[width=0.49\textwidth]{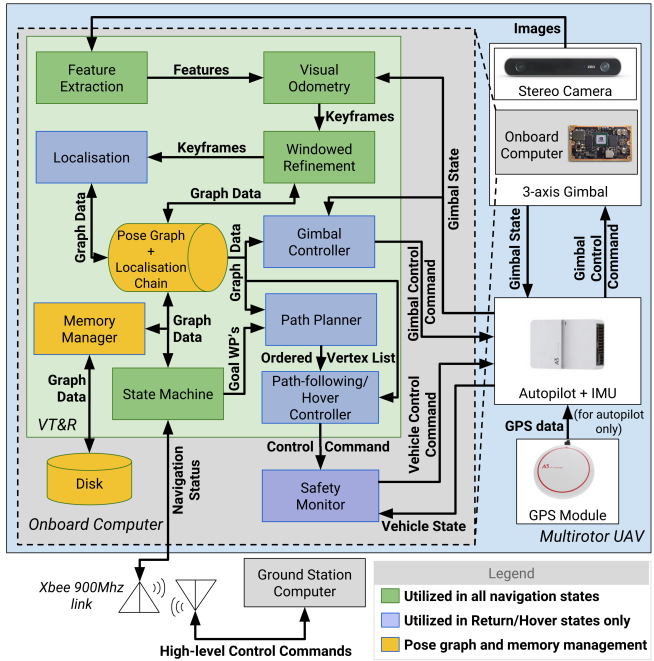}
  \caption{The architecture of the \ac{vtr} system for multirotor \acp{uav}.}
  \label{fig:arch}
  \vspace{-7mm}
\end{figure}

The architecture of the \ac{vtr} system for the multirotor \ac{uav} is shown in Fig. \ref{fig:arch}. All processing, including visual navigation, localisation, planning and control occurs on-board the \ac{uav} on the primary computer (Fig. \ref{fig:m600}). This computer directly interfaces with the on-board camera via \ac{USB} 3.0, which provides grayscale stereo images for visual navigation. This computer also interfaces with the on-board autopilot via a serial \ac{TTL} connection, which provides vehicle data (gimbal state, autopilot state, etc.) and the interface for sending control commands. A long-range, low-bandwidth 900 Mhz wireless link is used to communicate with the primary on-board computer from a ground station. The ground station computer is utilized only for status monitoring and sending of high-level control commands. These commands consist of manual state transition requests (switching from \textit{learn} to \textit{return}), obtaining flight control authority from the autopilot, and initiating \ac{GPS} waypoint missions.

The \ac{vtr} software system consists of several interacting components (Fig. \ref{fig:arch}): 1) \ac{VO}, 2) windowed refinement, 3) visual localisation, 4) a state machine, 5) gimbal and path-following controllers and 6) a safety monitor. Each system operates in a separate thread or process, interacting through the transfer of data caches (a packet of new and derived data, including images, processed features and estimated transforms) and through the use of a Google Protobuf backend for disk storage. Memory managers ensure that stale data is written to disk to reduce \ac{RAM} utilisation, and is pre-emptively re-loaded during the return phase to ensure localisation can proceed without waiting for disk access. The \ac{ROS} is used to run the safety monitor and interface to the autopilot and camera. The adapted \ac{vtr} state machine for multirotor \acp{uav} controls the high-level state that the system is in (usually \textit{learn} or \textit{return}).

A safety monitor runs as an independent process to ensure safe operation of the vehicle in case of system failure. It performs a sanity check control and localisation data, in addition to a watchdog functionality on the control commands and state data both from \ac{vtr} and the autopilot. Any monitored command or state data that is delayed by more than a preconfigured timeout triggers a safety failure, forcing the vehicle to release software control and revert to manual pilot control.

In the following sections, the visual system, path-following and hover controllers are described in more detail.

\subsection{Visual System}
The visual system consists of seperate threads for feature extraction, pose estimation (\ac{VO}), refinement and localisation, using images captured by a stereo camera to estimate both pose updates and localisation to the path during the return.
\subsubsection{Visual Odometry}
During both the \textit{learn} and \textit{return} phases, image pairs are captured by a calibrated stereo camera at a frame rate of $\sim$15 Hz, while the gimbal state (read as roll-, pitch-, and yaw-axis angular positions) is captured at 10 Hz. The gimbal state gives the pose of the camera in the vehicle frame by compounding the captured gimbal angles through a series of transforms with known translations extracted from 3D vehicle models. We denote the vehicle-to-sensor (camera) transform at time $\tau$ as $\mathbf{T}_{sv}^{\tau}$. 

For each stereo image pair captured at time $t$, \ac{SURF} features are extracted, descriptors generated and landmarks triangulated. Landmarks are triangulated from both the stereo pair and from motion to account for both close proximity and extremely large depths depending on altitude, similar to \cite{Mur-Artal2017}. Each feature in this latest frame-pair is matched to the last keyframe via \ac{SURF} descriptor matching on the \ac{GPU}. The raw matches are then passed through a \ac{MLESAC} robust estimator to find the relative transform to the last keyframe. Finally, this transform is optimised using our \ac{STEAM} bundle adjustment engine \cite{7353368}, keeping landmarks fixed.

After this process, if the number of inliers drops below a minimum count or the motion (translation or rotation) exceeds a threshold, the frame is set as a keyframe and the features, new landmarks, and vehicle-to-sensor transform at that time are stored in a vertex in a pose graph for future retrieval. The relative transform is stored as an edge to the previous vertex. Windowed bundle adjustment (termed \textit{windowed refinement}) is then performed on the last 5-10 vertices. This \ac{VO} plus bundle adjustment process generates a dead-reckoned set of linked poses that represent the path. During the \textit{learn} phase, this set of poses and edges is marked as `privileged'. Naturally, incremental translational and rotational errors compound during this process, causing the global map to be distorted. However, \ac{vtr} depends on the graph being only \textit{locally} metric in the region to which the vehicle is localized. For a more thorough explanation of this component, we direct the reader to our previous work~\cite{Paton2016}.
\subsubsection{Visual Localisation}
\begin{figure}[tbp]
\vspace{-3mm}
  \centering
  \vspace{2mm}
  \includegraphics[width=0.49\textwidth]{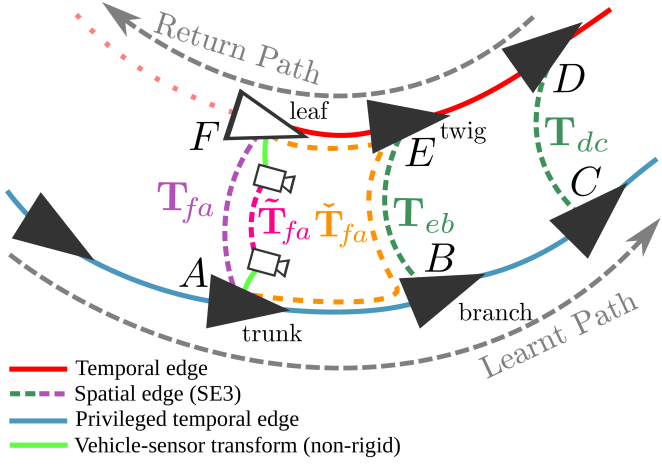}
  \caption{During the return phase, the vehicle follows the learned route in reverse. The localisation chain updates the estimated localisation transform $\mathbf{\tilde{T}}_{fa}$ at each VO update. Upon creation of a new vertex $F$, visual localisation inserts the new edge $\mathbf{T}_{fa}$. The gimbal controller minimises orientation error of $\mathbf{\tilde{T}}_{fa}$, which includes vehicle-to-sensor transform, $\mathbf{T}_{sv}$, and $\mathbf{\check{T}}_{fa}$. The uncertainties and some estimated transforms are omitted here for clarity.}
  \label{fig:graph_overview}
  \vspace*{-5mm}
\end{figure}

During the \textit{return} phase, while the vehicle flies in reverse, an additional thread performs visual matching to the local map of 3D points in the graph to estimate the path-following error (Fig. \ref{fig:graph_overview}), which is used by a path-following controller to keep the vehicle on the path. 

To enable this process, the \textit{localisation chain} is used to keep track of important vertices in the graph and their respective transforms. We use a `tree' model to name vertices in the chain, going from the \textit{trunk} vertex (defined as the closest vertex spatially on the privileged path), through the \textit{branch} (the closest vertex on the privileged path with a successfully \ac{MLESAC} estimated transform), \textit{twig} (the corresponding vertex on the current path) and \textit{leaf} (latest live vertex) vertices. These can be seen in Fig. \ref{fig:graph_overview}. We use the notation $\mathfrak{t},\mathfrak{b},\mathfrak{w}$ and $\mathfrak{l}$ to refer to the \textit{trunk}, \textit{branch}, \textit{twig} and \textit{leaf} vertices, respectively.

At every step of \ac{VO} (i.e., on every successfully estimated frame, not just keyframes) the localisation chain is updated with the estimated transform from \textit{trunk} to \textit{leaf}, (or $\mathbf{\check{T}}_{\mathfrak{lt}} = \mathbf{\check{T}}_{fa} = \mathbf{T}_{fe}\mathbf{T}_{eb}\mathbf{T}_{ba}$ in Fig. \ref{fig:graph_overview}). The leaf is updated every step and, if necessary, the \textit{trunk} vertex is updated to the closest estimated privileged vertex to the \textit{leaf}.

Upon insertion of a new \ac{VO} keyframe as a vertex in the graph, the localisation thread attempts to estimate a new transform from \textit{branch} to \textit{twig}. This process follows four separate stages: i) landmark migration, ii) landmark matching, iii) pose estimation, iv) optimisation. First, the nearest privileged vertex (the \textit{trunk}) is used as the base vertex to generate a local window of privileged vertices that contain potentially matchable landmarks. Using the transforms on the privileged edges, the landmarks in this window are transformed to the \textit{trunk} to generate a locally metric set of 3D points with a common origin\footnote{While our previous work incorporates features and points from multiple experiences (i.e., multiple traverses), the learn-return framework by definition only uses a single experience: the privileged one.}.

Following a similar process to \ac{VO}, features in the latest non-privileged vertex (the \textit{leaf}) are matched using their \ac{SURF} descriptors to \textit{all} descriptors of the migrated landmarks, which are then passed through a \ac{MLESAC} robust estimator to estimate the relative transform from \textit{trunk} to \textit{leaf}, $\mathbf{T}_{\mathfrak{lt}}$ ($\mathbf{T}_{fa}$ in Fig. \ref{fig:graph_overview}). Finally, the transform is optimised while leaving all landmarks fixed. The localisation chain is then updated to reflect this fresh transform estimate, and the new \textit{branch} to \textit{twig} is set $\mathbf{T}_{\mathfrak{wb}}^\ast \leftarrow \mathbf{T}_{fa}$. The path-following controller can query the localisation chain at any time to get the best estimate of $\mathbf{T}_{\mathfrak{lt}}$, facilitating control at high speed even with significant delays from visual localisation.

\subsection{Gimbal Controller}
Use of a gimbal decouples the visual perspective from the roll/pitch-to-move actuation of multirotor \acp{uav}. This significantly improves the robustness of \ac{vtr} in the air by adding extra degrees of actuation to the visual servoing problem. During fast, dynamic maneuvers, a gimballed camera system will be able to outperform a static camera system by decoupling the aircraft motion from the camera view. In addition, maintaining a consistent roll ensures that generally unstable point features are tracked more consistently.

During the \textit{learn} phase, gimbal control is not performed by \ac{vtr}, but left open-loop such that the gimbal internal controller performs stabilisation of roll and pitch, and smoothes yaw that follows the vehicle yaw. The value of this sensor-to-vehicle transform $\mathbf{T}_{sv}^\tau$ (Fig. \ref{fig:graph_overview}) is recorded at each new vertex, corresponding to time $\tau$. During the return, the gimbal is actively controlled by \ac{vtr} for the pitch and yaw axes. The gimbal is commanded to reduce orientation error between the current (\textit{leaf}) view and nearest privileged (\textit{trunk}) view, ($\mathbf{\tilde{T}}_{fa}$ in Fig. \ref{fig:graph_overview}), as knowledge of the transform between the current and the privileged poses is known via the localisation chain such that:
\begin{equation}
\mathbf{\tilde{T}}_{fa} = \mathbf{T}_{sv}^\mathfrak{l}\mathbf{T}_{\mathfrak{lt}}\inv{\mathbf{T}_{sv}^\mathfrak{t}}
\end{equation}
using the sensor-to-vehicle transforms captured at vertices $\mathfrak{t}$ and $\mathfrak{l}$. $\mathbf{\check{T}}_{fa}$ is updated in the localisation chain at every frame.
\subsection{Path-Following controller}
A path-following controller is implemented for vehicle control during the \textit{return} phase to keep the vehicle as close as possible to the outbound path while mainitaining a suitable target velocity.

To enhance robustness to environmental disturbances and system delays, we consider a path-following approach, which, in contrast to trajectory tracking, prioritizes spatial error over temporal error \cite{Hauser1995}. By extending the approach in \cite{Hauser1995} to a \ac{vtr} framework, we achieve simple multirotor path-following. This is done by converting a standard P-D tracking control to select the spatially closest reference point on the path at each control time step (50 Hz). 

\begin{equation*}
	\mathbf{\check{T}}_{\mathfrak{tl}} = \begin{bmatrix}
	\mathbf{C}_{\mathfrak{tl}} & \mathbf{p}^{\mathfrak{lt}}_\mathfrak{t} \\ \mathbf{0}^T & 1
	\end{bmatrix} = \mathbf{\check{T}}^{-1}_{\mathfrak{lt}} . 
\end{equation*}
\noindent
We obtain a translational velocity estimate $\mathbf{v}^{\mathfrak{lt}}_\mathfrak{t} = (\dot{x}, \dot{y}, \dot{z})$ using STEAM trajectory generation \cite{7353368}, which fits a constant velocity trajectory through the previous path vertexes. 

\paragraph{\ac{vtr} Path-Following Reference} We generate a path by connecting a straight-line through successive privileged vertices. To do this, we use the localization chain to obtain a transform from the \textit{next} privileged vertex to the \textit{trunk} $\mathbf{T}_{\mathfrak{tn}}$. From this we can extract the position $\mathbf{p}^{\mathfrak{nt}}_\mathfrak{t}$ of the \textit{next} privileged vertex with respect to the \textit{trunk} using
\begin{equation*}
	\mathbf{T}_{\mathfrak{tn}} =  \begin{bmatrix}
	\mathbf{C}_{\mathfrak{tn}} & \mathbf{p}^{\mathfrak{nt}}_\mathfrak{t} \\ \mathbf{0}^T & 1
	\end{bmatrix}.
\end{equation*}

\noindent
At each time step, we determine the reference position $\mathbf{p}_{\rm ref} = (x_{\rm ref}, y_{\rm ref}, z_{\rm ref})$ by projecting our current multi-rotor position $\mathbf{p}^{\mathfrak{lt}}_\mathfrak{t}$ onto the straight-line segment connecting the \textit{trunk} to the \textit{next} privileged vertex using:
\begin{equation*}
	\mathbf{p}_{\rm ref} = \mathbf{p}^{\mathfrak{lt}}_\mathfrak{t} \cdot \mathbf{p}^{\mathfrak{nt}}_\mathfrak{t} \frac{\mathbf{p}^{\mathfrak{nt}}_\mathfrak{t}}{|\mathbf{p}^{\mathfrak{nt}}_\mathfrak{t}|}.
\end{equation*}
\noindent
We obtain a reference velocity $\mathbf{v}_{ref}= (\dot{x}_{ref}, \dot{y}_{ref}, \dot{z}_{ref})$, where the magnitude is a user-selected parameter $v_{des}$, in the direction of the \textit{next} privileged vertex using:
\begin{equation*}
	\mathbf{v}_{\rm ref} = v_{des} \frac{\mathbf{p}^{\mathfrak{nt}}_\mathfrak{t}}{|\mathbf{p}^{\mathfrak{nt}}_\mathfrak{t}|}.
\end{equation*}

\paragraph{Control Design} Our path-following control is designed to send commands $(\dot{z}_{\rm cmd}, \dot{\psi}_{\rm cmd}, \theta_{\rm cmd}, \phi_{\rm cmd})$ where $\dot{z}_{\rm cmd}$ is a commanded $z$-velocity, $\dot{\psi}_{cmd}$ is a command yaw rate, and $\theta_{\rm cmd}$ and $\phi_{\rm cmd}$  are commanded pitch and roll, respectively. The $z$-velocity command is designed using a P-D controller:  
\begin{equation}
\label{eq_z_cmd}
	\dot{z}_{\rm cmd} = \frac{2 \zeta_z}{\tau_z} (z_{\rm ref} - z) + \frac{1}{\tau_z^2} (\dot{z}_{\rm ref} - \dot{z}),
\end{equation}

\noindent
where $\zeta_z$ and $\tau_z$ are tuned damping ratio and time constant. The current yaw, $\psi$, with respect to the \textit{trunk} is determined from the rotation matrix, $\mathbf{C}_{\mathfrak{tl}}$. As seen in (\ref{eq_yaw_cmd}), a P-controller (with tuned time constant $\tau_{\psi}$) is used to correct for any yaw-mismatch between the \textit{leaf} and the \textit{trunk}: 
\begin{equation}
\label{eq_yaw_cmd}
	\dot{\psi}_{cmd} = -\frac{1}{\tau_{\psi}} \psi .
\end{equation}

As in \cite{Spedicato2016}, lateral-motion control commands are determined by first designing translational acceleration commands using P-D control:
\begin{subequations}
\begin{equation}
	a_x = \frac{2 \zeta_{\theta}}{\tau_{\theta}} (x_{ref} - x) + \frac{1}{\tau_{\theta}^2} (\dot{x}_{ref} - \dot{x}),
\end{equation}
\begin{equation}
	a_y = \frac{2 \zeta_{\theta}}{\tau_{\theta}} (y_{ref} - y) + \frac{1}{\tau_{\theta}^2} (\dot{y}_{ref}- \dot{y}),
\end{equation}
\end{subequations}
\noindent
where $\zeta_\theta$ and $\tau_\theta$ are tuned damping ratio and time constant. Assuming small lateral acceleration ($\ddot{x} \approx \ddot{y} \approx 0$) and using standard feedback linearization, these linear acceleration commands are transformed into pitch and roll commands:
\begin{subequations}
\begin{equation}
	\theta_{cmd} = \arcsin ( \frac{a_x}{g} \cos\psi +  \frac{a_y}{g} \sin\psi ),
\end{equation}
\begin{equation}
	\phi_{cmd} = -\arcsin ( -\frac{a_x}{g} \sin\psi +  \frac{a_y}{g} \cos\psi ),
\end{equation}
\end{subequations}
\noindent
where $g$ is the gravitational constant.

\section{EXPERIMENTS}\label{sec:experiments}
To evaluate the performance of the airborne \ac{vtr} algorithm, a number of outdoor experiments were performed on-board the target \ac{uav}. 

In the first experiment, we evaluate the performance of the localisation algorithm under \ac{GPS} control using the described gimbal controller. Specifically, we test the performance of the localisation algorithm and gimbal controller under deliberately challenging conditions, including high-speed, dynamic flight and high \textit{learn} vs \textit{return} positional error. For this experiment, we deliberately exclude the vehicle controller to isolate the performance of the subcomponents of the algorithm. In the second experiment, we perform closed-loop control with the aforementioned path-following controller developed for full 6-DOF vehicle motion. This system is evaluated over several runs, showing the full system operating. The experimental setup is described in the following subsection.

\subsection{Experimental Setup}
For these experiments, we use a DJI Matrice 600 Pro, with attached Ronin-MX gimbal (Fig. \ref{fig:m600}). This system has a take-off weight of approximately 10kg, and maximum span rotor-tip-to-tip of 1.64m. Control is provided by a DJI A3 triple redundant autopilot. On-board this system is an NVIDIA Tegra TX2 module (6 ARM cores + 256 core Pascal GPU) and StereoLabs ZED stereo camera connected via USB, both mounted in the stabilised platform of the Ronin-MX gimbal. The Matrice 600 Pro provides state information to \ac{vtr} running on-board the TX2, including gimbal encoder positions and \ac{GPS} status, while the ZED camera provides grayscale imagery with resolution $672\times376$ at 15 Hz. The Tegra TX2 runs NVIDIA L4T v28.2, a variant of Ubuntu 16.04 for ARM architectures.

The primary location used for the experiments in this paper is a simulated village at the \ac{DRDC} Suffield Research Centre in southern Alberta, Canada. The Suffield location consists of a number of shipping containers placed to emulate buildings and narrow alleys in flat grassland, suited to a simulated patrol scenario.

\begin{figure}
  \centering
  \includegraphics[width=0.49\textwidth]{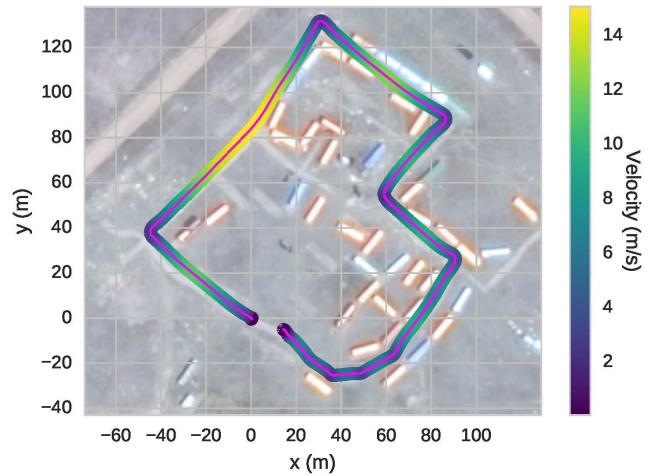}
  \vspace{-6mm}
  \caption{Overview of the trajectory flown at the \ac{DRDC} Suffield Research Centre, shown in magenta. Velocity profile for a target 15 m/s commanded speed overlaid.}
  \label{fig:suffield_map}
  \vspace{-7mm}
\end{figure}

\subsection{Localization Performance Evaluation}
In these experiments, we evaluate the combined performance of the localisation algorithm and gimbal controller to successfully localise the vehicle under increasingly difficult operational conditons. We test this in two ways: increasing target velocity of the vehicle, and deliberately offset altitudes on the outbound and return paths. The first test shows the performance under increasingly dynamic maneuvers of the vehicle, inducing rapid perspective change and poor path tracking, which must be attenuated by the gimbal controller. The second test shows the performance of the localisation algorithm with intentionally poor perspective. We deliberately do not use the vehicle controller in these tests to decouple and isolate the performance of the localisation algorithm and gimbal controller.


\subsubsection{Increasing Target Velocity}
For this experiment, the aircraft is autonomously flown at 12m \ac{AGL} along the path depicted in Fig. \ref{fig:suffield_map} in a clockwise direction. \ac{vtr} is placed into \textit{learn} mode, before the outbound route is flown under autonomous control, by uploading a waypoint mission to the Matrice 600 autopilot. Once the vehicle reaches the end of the loop, \ac{vtr} is switched to \textit{return} mode, and the aircraft is again autonomously commanded to return along the same path by following the waypoints in reverse. During this return stage, the gimbal is actively controlled by \ac{vtr} to reduce orientation errors caused by path-following discrepencies generated by the GPS-based controller.

The route is flown at increasingly fast target speeds, 3, 7, 8, 10, 12 and 15m/s, on both the learn and return stages. While the vehicle reaches this speed during only parts of this path, the average speed also increases with each pass. A typical speed profile for the path at 15m/s target speed is shown in Fig. \ref{fig:suffield_map}.

\begin{figure}
\includegraphics[width=0.49\textwidth]{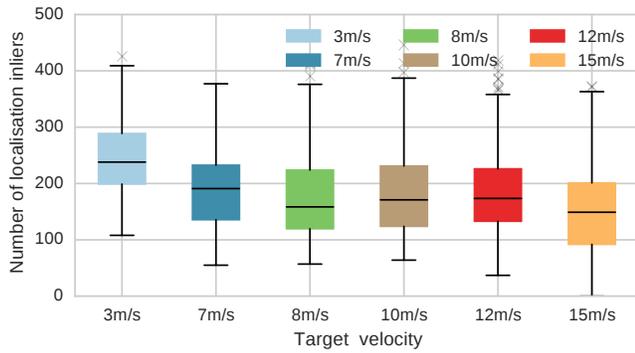}
\vspace{-6mm}
\caption{Localisation performance is comparable with increasing target (and average) velocity, where \textit{learn} and \textit{return} phases are conducted at the same speed.}
\label{fig:loc_inliers_speed}
\vspace{-7mm}
\end{figure}

Fig. \ref{fig:loc_inliers_speed} shows the median and variance of the localisation inliers recorded along the return path for each of the target speeds. This figure shows that even at the highest commanded speed (15m/s), localization performance remains similar to those examples at lower speeds. At 15m/s some localisation failures occur, but the majority of these can be attributed to failures of the hardware gimbal controller during a segment of the path.


\subsubsection{Increasing Height Error}
\begin{figure}
\includegraphics[width=0.49\textwidth]{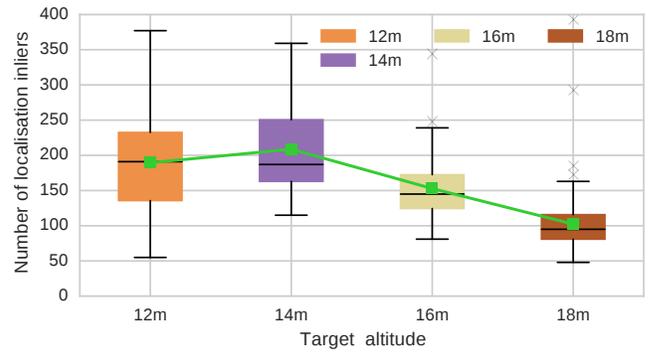}
\vspace{-6mm}
\caption{Successful localisation occurs with significantly increasing altitude difference between \textit{learn} (12 m) and \textit{return} phases (tested at 12, 14, 16 and 18 m, but the average (green) shows decline at more extreme (50\%) differences.}
\label{fig:loc_inliers_height}
\vspace{-5mm}
\end{figure}

For this experiment, the aircraft is again autonomously flown at 12m \ac{AGL} during the \textit{learn} stage along the path depicted in Fig. \ref{fig:suffield_map} in a clockwise direction at a nominal speed of 7 m/s. The total length of the path is approx. 450m. Once the vehicle reaches the end of the loop, \ac{vtr} is switched to \textit{return} mode, and the aircraft is again autonomously commanded to return along the same path by following the waypoints in reverse at the same 7 m/s target speed. In this case, however, we vary the altitude at which the aircraft returns, to test the robustness of the gimbal controller and ability of the algorithm with large positional offsets. In these experiments, we show the localisation inliers along the path with target return heights of 12, 14, 16 and 18m, respectively (Fig. \ref{fig:loc_inliers_height}).


In Fig. \ref{fig:loc_inliers_height}, localisation performance is still high, with an average of 100 inliers per keyframe, even at altitude differences of 6m, or 50\%. While some of this performance can be attributed to perspective due to the altitude, a significant component can be attributed to the gimbal compensating for the reduced image overlap that would be present on a static camera. Importantly, however, the average inliers does drop significantly, and more interestingly, reduces in variance. This is likely due to the enhanced viewpoint overlap (of the \textit{learnt} path) at higher altitudes, meaning positional errors have less effect on maintaining observability of all landmarks during localisation.


\begin{figure}
\includegraphics[width=0.49\textwidth]{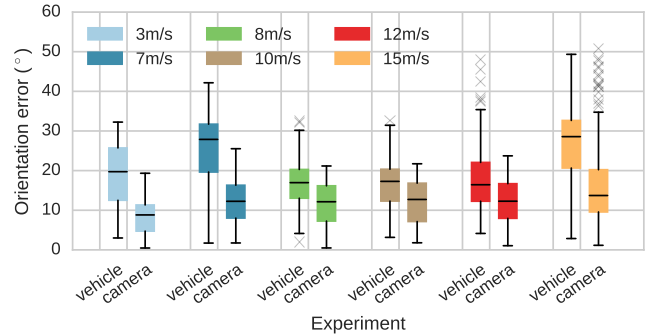}
\vspace{-6mm}
\caption{While vehicle attitude error increases in both median and variance with increasing target velocity, the gimbal controller maintains a consistent camera orientation between \textit{learn} and \textit{return} regardless of target speed.}
\label{fig:ori_error}
\vspace{-7mm}
\end{figure}

Finally, Fig. \ref{fig:ori_error} shows the utility of the gimbal in minimising perspective error caused by differing vehicle attitudes between \textit{learn} and \textit{return}. Due to the pitch-to-move nature of multirotor systems, accelerations and decelerations cause the vehicle attitude to differ between these two passes of the path. For a static camera system, these differences can cause performance degradation due to poor image overlap. Using a gimbal with active control to minimise camera orientation can minimise this effect. Fig. \ref{fig:ori_error} shows the magnitude of orientation error for two separate localisation transforms at each speed profile (read as a pair) taken from the estimated localisation chain as estimated from the visual pipeline: in the vehicle frame ($\mathbf{T}_{\mathfrak{lt}}$, or $\mathbf{T}_{fa}$ in Fig. \ref{fig:graph_overview}) on the left, and in the camera frame ($\mathbf{\tilde{T}}_{fa}$ in Fig. \ref{fig:graph_overview}) on the right. As can be seen at all speed profiles, the gimbal succeeds in minimising the orientation of the localisation transform, and this performance is relatively consistent with increasing speed. In this scenario, the target speeds of the \textit{learn} and \textit{return} phases are the same for each speed profile, meaning there will be some consistency in orientation in both phases. With differing speed profiles, we would expect the observed utility of the gimballed camera to increase further.

\subsubsection{Execution Time}

\begin{figure*}
  \centering
  \includegraphics[width=0.99\textwidth]{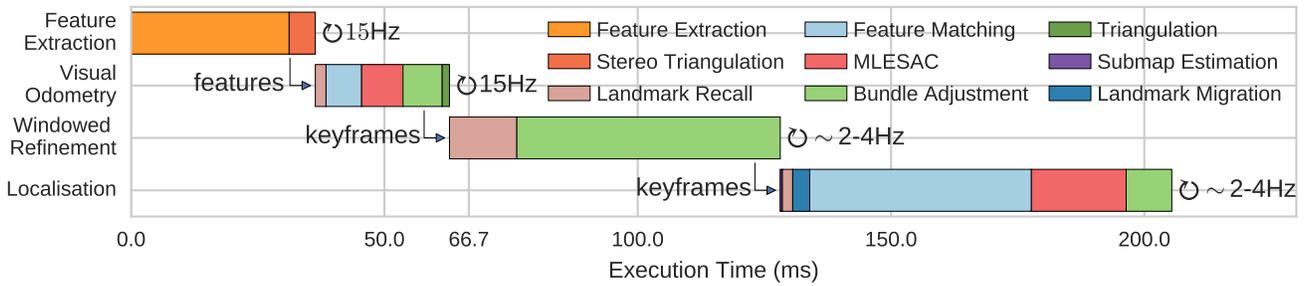}
  \vspace{-3mm}
  \caption{Average execution times for the visual pipelines on the TX2 during a misison, seperated by thread of execution. Once one thread is finished processing, it is able to process the next image and its data products.}
  \label{fig:execution_times}
  \vspace{-7mm}
\end{figure*}

Fig. \ref{fig:execution_times} shows the average execution time for the seperate processes in the \ac{vtr} software on-board the Tegra TX2. While \textit{feature extraction} and \textit{\ac{VO}} process every image pair at an approximate speed of 66ms ($\sim$15 Hz), \textit{windowed refinement} only runs on generation of a keyframe, and \textit{localisation} runs after this process is complete. Feature extraction and matching are all performed on the GPU. Using this threaded setup allows \ac{vtr} to run online.

\subsection{Full VT\&R Evaluation}
In this experiment, we evaluate the performance of the full closed-loop \ac{vtr} system, using \ac{GPS} navigation during the \textit{learn} phase, and switching to the presented path-following controller for the \textit{return} phase.  Over three separate trials, each consisting of a single flight, we traverse the path shown in Fig. \ref{fig:suffield_map} in a clockwise direction at an altitude of 12m \ac{AGL}, before returning in an anticlockwise direction at the same altitude (attempting to minimise all positional errors) at a target speed of 3m/s. 

\begin{figure}
  \centering
  \includegraphics[width=0.49\textwidth]{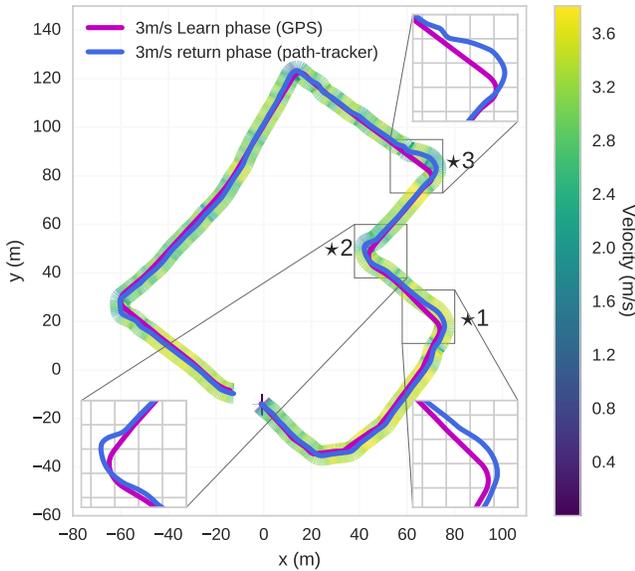}
  \vspace{-8mm}
  \caption{The outbound (GPS, magenta) and return (controller, blue) paths during a single trial. For the majority of the path, the controller maintains less than 1.5m cross-track (y-z) error. Some offset is seen on sharp turns.}
  \label{fig:control_profile}
  \vspace{-7mm}
\end{figure}

\begin{figure}
  \centering
  \includegraphics[width=0.49\textwidth]{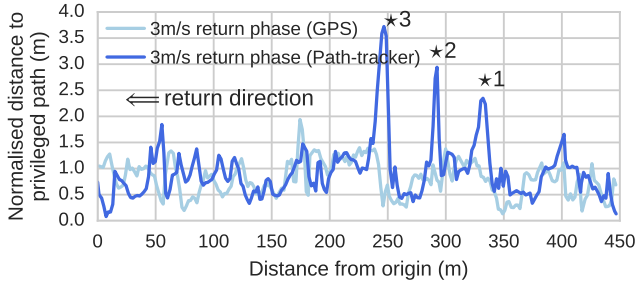}
  \vspace{-7mm}
  \caption{Path-following error is of a similar order to that for GPS over the majority of the path using our controller, according to the localisation transform estimated by \ac{vtr}.}
  \label{fig:dist_control}
  \vspace{-3mm}
\end{figure}

\begin{figure}
  \centering
  \includegraphics[width=0.49\textwidth]{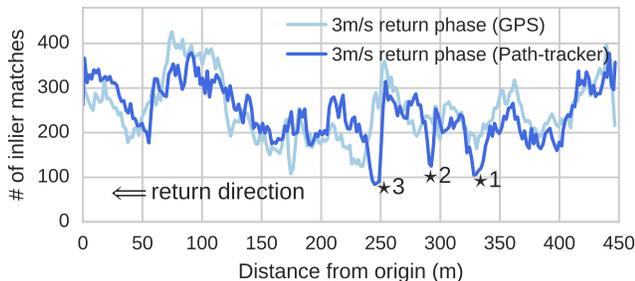}
  \vspace{-7mm}
  \caption{The number of localization inliers while using our path-following controller is of a similar order to that for GPS over the majority of the path.}
  \label{fig:inliers_control}
  \vspace{-7mm}
\end{figure}

In all three trials, \ac{vtr} was able to complete the return phase of flight under path-following control over an approximately 2 minute period. Fig. \ref{fig:control_profile} shows the path for one of these trials. The outbound path under GPS control is shown in magenta, while the return path under path-following control is shown in blue. Figs. \ref{fig:dist_control} and \ref{fig:inliers_control} show the normalised cross-track error (in Y and Z, using the vision-based estimate) and number of inlier matches respectively. Specific segments of the path are highlighted in the inset figures of Fig. \ref{fig:control_profile} and annotated with numbers that correlate to those in Figs. \ref{fig:dist_control}-\ref{fig:inliers_control}.

The positional error is less than 1.5m over most of the path using the path-following controller, and is comparable to a return trajectory under \ac{GPS} control, showing the strong performance of a simple vision-based path-following controller compared to this primary sensor. In specific sections such as corners, however, cross-track error increases to a maximum of 3.6 m. This can be attributed to the simplicity of the controller, as curvature of the path is not accounted for, and velocity error is weighted higher than cross-track error.

Additonally, localisation performance is strong over the full trajectory, with no localisation failures, even at the highlighted corner points. The average performance over the trajectory is again comparable to a return phase under GPS control.



\vspace{-2mm}
\section{CONCLUSIONS}\label{sec:conclusion}
In this paper, a full \ac{vtr} system for emergency return of a multirotor \ac{uav} has been presented. Using 15 Hz imagery from a gimbal-stabilised stereo camera to build a map online during a commanded \textit{learn} phase, we have demonstrated autonomous return of the vehicle by matching landmarks back to a live view for autonomous path-following control with equivalent path-following errors to the on-board GPS system. In addition, we have demonstrated the robustness of the gimbal-stabilised system to high-speeds and large positional errors.

Future work will include the development of a more advanced path-tracking controller that uses path curvature to minimise cross-track errors, and testing in a multi-experience framework over a long-term experiment.


\section*{ACKNOWLEDGMENT}
This work was funded by Smart Computing for Innovation Consortium (SOSCIP), Defense Research and Development Canada (DRDC), Drone Delivery Canada (DDC), Natural Sciences and Engineering Research Council of Canada (NSERC) and the Centre for Aerial Robotics Research and Education (CARRE), University of Toronto.
\bibliographystyle{ieeetr}
\vspace{-1mm}
\bibliography{bib/refs}
\end{document}